\documentclass{article}
\usepackage{ijcai23}

\usepackage{times}
\usepackage{soul}
\usepackage{url}
\usepackage[hidelinks]{hyperref}
\usepackage[utf8]{inputenc}
\usepackage[small]{caption}
\usepackage{graphicx}
\usepackage{amsmath}
\usepackage{amsthm}
\usepackage{booktabs}

\usepackage{subcaption}
\usepackage[switch]{lineno}
\usepackage{algorithm}
\usepackage[noend]{algpseudocode} %

\title{Enhancing Accuracy and Robustness through Adeversarial Training\\in Class Incremental Continual Learning}

\author{
    Minchan Kwon$^1$
\and
Kangil Kim$^2$\footnote{corresponding author}
\affiliations
$^1$    Graduate School of AI, KAIST, Deajeon, Republic of Korea
    \emails
    kmc0207@kaist.ac.kr
\\
$^2$AI Graduate School, GIST, Gwangju, Republic of Korea, 
\emails
kangil.kim.01@gmail.com
}

\begin{document}

\maketitle

\begin{abstract}
    In real life, adversarial attack to deep learning models is a fatal security issue. However, the issue has been rarely discussed in a widely used class-incremental continual learning (CICL). 
    In this paper, we address problems of applying adversarial training to CICL, which is well-known defense method against adversarial attack. 
    A well-known problem of CICL is class-imbalance that biases a model to the current task by a few samples of previous tasks.
    Meeting with the adversarial training, the imbalance causes another imbalance of attack trials over tasks. 
    Lacking clean data of a minority class by the class-imbalance and increasing of attack trials from a majority class by the secondary imbalance, adversarial training distorts optimal decision boundaries. The distortion eventually decreases both accuracy and robustness than adversarial training. 
    To exclude the effects, we propose a straightforward but significantly effective method, \textit{External Adversarial Training} (EAT) which can be applied to methods using experience replay. 
    This method conduct adversarial training to an auxiliary external model for the current task data at each time step, and applies generated adversarial examples to train the target model. 
    We verify the effects on a toy problem and show significance on CICL benchmarks of image classification.
    We expect that the results will be used as the first baseline for robustness research of CICL. 
    
\end{abstract}

\section{Introduction}

Deep learning has achieved remarkable performance in various field of computer vision. However, it remains vulnerable to adversarial attacks, which add minuscule perturbations to an image that are almost imperceptible to the human eye but cause the model to make incorrect predictions. This has made adversarial attacks a major concern for researcher, as they pose a significant security risk when deep learning is applied in real-world scenarios. Therefore, developing defenses against and methods for launching adversarial attacks have become a focus of research in the field.

Despite the significance of continual learning (CL) in real-world applications of deep learning, there has been limited research on adversarial attacks and defenses in this context. CL examines how models can effectively learn from a stream of continuous data.
In our empirical analysis to reveal the impact of attacks, we found that class-incremental CL (CICL) setting is vulnerable to adversarial attack.
Furthermore, adversarial training (AT), the most widely used adversarial defense method, is ineffective in CICL settings. 
Compared to the expected robustness enhancing and small clean accuracy loss of AT in a single task, the AT in class-incremental CL shows larger drop of clean accuracy and only small improvement of robustness.
We argue that the cause of this problem is that the class imbalance, an inherent property of CL, deepens the model disturbance effect of AT.

To address these problems, we propose \textit{External Adversarial Training} (EAT), an adversarial training method that can create adversarial examples that exclude the class imbalance problem of CICL. EAT can be easily applied to any method using experience replay (ER)  which includes the SOTA models. To the best of our knowledge, EAT is the most effective method for defending against adversarial attacks while maintaining clean accuracy.
We verify and analyze the points on state-of-the-art and well-known rehearsal-based CICL methods on on split CIFAR-10 and split tiny-imagenet benchmarks. 

In summary, our contributions are as follows. 
\begin{itemize}
    \item verifying AT is ineffective in CICL
    \item analyzing the causes of the problem based on attack overwhelming
    \item presenting a simple but effective EAT method to exclude the causes 
    \item providing baseline of robustness for several rehearsal-based method 
\end{itemize}


\section{Background}
\paragraph{Class-Incremental Continual Learning}
Continual learning is environments in which a model called target model is trained to learn new tasks or classes in sequential manner, without forgetting the previously learned tasks or classes. This means that the model is continually exposed to new data and must learn to adapt to the new information while retaining the knowledge it has gained from previous tasks.
There are many different settings for continual learning, following recent CL literature~\cite{mai2022online,cha2021co2l,buzzega2020dark}, we consider the supervised class-incremental continual learning setting where a model needs to learn new classes continually without task-iD.
The stream $D$ is a sequence of disjoint subsets whose union is equal to the whole training data, notated as $\{T_1, \cdots, T_N\}$ where $T_i$ indicates the subset called a \textit{task} at $i$th time step. Each task is a set of input and ground truth label pairs.
Training in the class-incremental continual learning has two constraints: 1) a \textit{target model} which want to training for continuous dataset composed of an encoder and single-head classifier is shared overall tasks, and 2) the model learns from only a task at each time step without accessibility to the other tasks. 
The single-head classifier uses all classes in $D$, not restricted to the classes of a task, which is more challenging environment than the other settings using task-IDs or using different classifier for tasks. In CICL setting, the model suffer from class imbalance because the previous task data is inaccessible.

\paragraph{Rehearsal-based method}
Rehearsal-based methods, also known as replay-based methods, are a popular approach for addressing the issue of catastrophic forgetting in CL. These methods use a memory buffer composed of a small fraction of previous training samples to reduce forgetting of previously learned information. The most typical method in this category is Experience Replay (ER) ~\cite{mai2022online,chaudhry2019tiny}. ER updates the network with training batches consisting of samples from both new and previous classes. The ER approach is simple and effective, but it has some limitations such as requiring extra memory to store the replay buffer.

DER/DERpp~\cite{buzzega2020dark} improves the performance of ER by leveraging distillation loss. CO2L~\cite{cha2021co2l} improves performance through contrastive learning. These methods have been shown to be effective in reducing forgetting and improving performance in class-incremental CL scenarios.

\paragraph{Adversarial Attack}
\textit{Adversarial example/image}, were first introduced by~\cite{szegedy2013intriguing}. These examples are modified versions of clean images that are specifically designed to confuse deep neural networks. \textit{Adversarial attacks} are methods for creating adversarial examples. These attacks can be classified into  various categories based on their goals and specific techniques. In this paper, we will focus on white-box attacks, which assume knowledge of the model's parameters, structure, and gradients.

Fast Gradient Sign Method (FGSM)~\cite{szegedy2013intriguing,goodfellow2014explaining} is a popular white-box attack that utilizes gradient information to update the adversarial example in a single step, in the direction of maximum classification loss. The FGSM update rule is given by $x' = clip_{[0,1]}\{x+\epsilon \cdot sign(\nabla_x,\mathcal{L}(x, y ; \theta))\}$. 
Basic Iterative Method (BIM)~\cite{kurakin2018adversarial} is an extension of FGSM which use iterative method to generate adversarial examples through multiple updates. Projected Gradient Descent (PGD) is similar to BIM, but with the added feature of randomly selecting an initial point in the neighborhood of the benign examples as the starting point of the iterative attack. PGD can be interpreted as an iterative algorithm to solve the following problem : $max_{x':||x'-x||_{\infty}<\alpha} \mathcal{L}(x', y ; \theta)$. PGD is recognized by ~\cite{athalye2018obfuscated} to be one of the most powerful first-order attacks. The use of random noise was first studied by ~\cite{tramer2017ensemble}. In the PGD attack, the number of iteration $K$ is crucial factor in determining the strength of the attacks, as well as the computation time for generating adversarial examples. In this paper, we will refer to a $K$-step PGD attack as PGD-$K$.

Adversarial defense methods have been widely studied in recent years due to the increasing concern for the security of deep learning models. These methods aim to improve the robustness of deep neural networks against adversarial attacks, which are specifically designed to exploit the weaknesses of the model by introducing small, imperceptible perturbations to the input data. \textit{Adversarial training} (AT)~\cite{goodfellow2014explaining} is a popular method that trains the model with generated adversarial examples, making the model more robust against similar attacks. \textit{Robustness} is used as a measure of how well the model defends an attack. This is a count of how much it is correct after applying an adversarial attack to the clean test data. To avoid confusion, in this paper, accuracy means clean accuracy using clean test data, and robustness means accuracy for adversarial attacks on clean test data.


\section{Problem of AT in CICL}

\begin{figure}[h]

  \centering
    \includegraphics[width=0.45\textwidth]{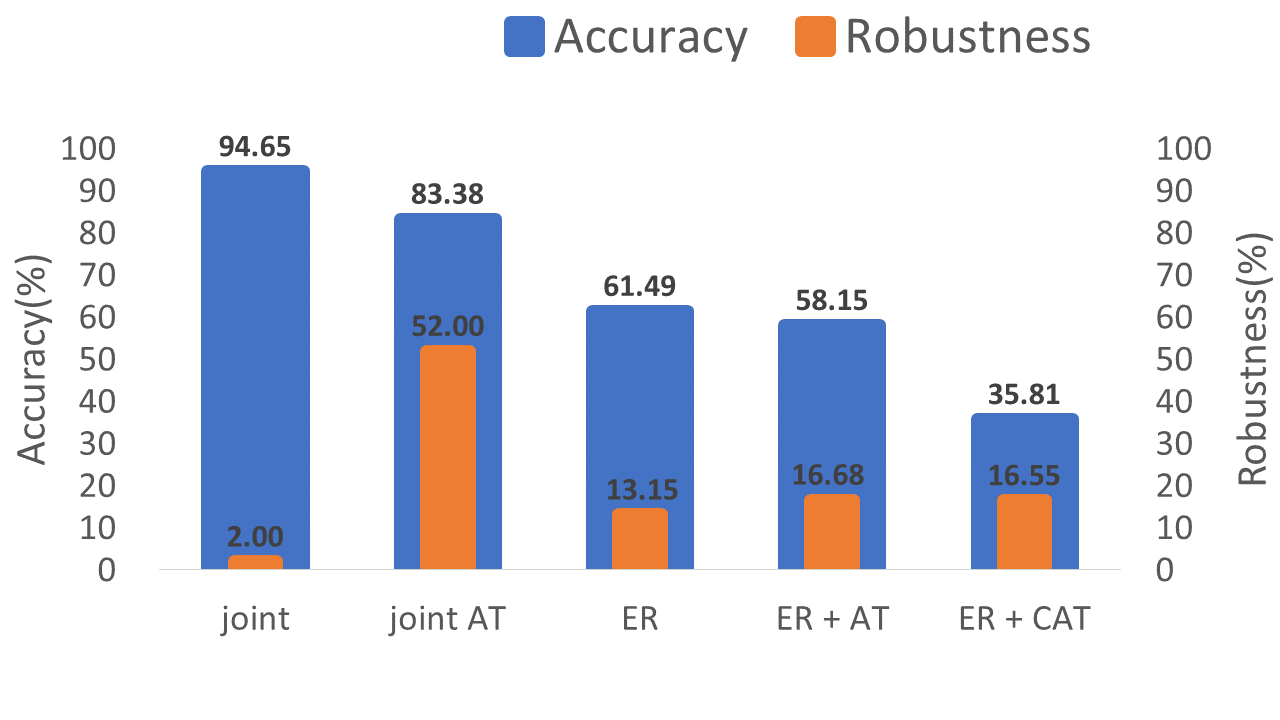}
  \caption{Accuracy and robustness of simple methods to apply AT to CICL with ER settings on split CIFAR-10 task.}
\label{fig:drawback_AT}
\end{figure}

\paragraph{Critical Drawback of Adversarial Training in CICL}

Naive application of AT to CICL causes serious problems on both robustness and accuracy. 
Figure~\ref{fig:drawback_AT} shows the negative impact of AT on a CICL data. This experiments conducted on sequencial CIFAR-10 and detail setting same as Section~\ref{setting}.
In the figure, applying AT to joint training decreases clean accuracy slightly, but increases the robustness dramatically. This is well known effect of AT~\cite{goodfellow2014explaining,zhang2019theoretically}.
However, AT in ER shows largely different results to this well-known effect of AT.
Clean accuracy significantly decreases and robustness also drops than joint adversarial training. 
This example shows the potential risk of AT in CICL framework.

\begin{figure*}[ht]
\begin{tabular}{cccc}

     \begin{subfigure}{0.223\textwidth}
         \includegraphics[width=1.2\linewidth]{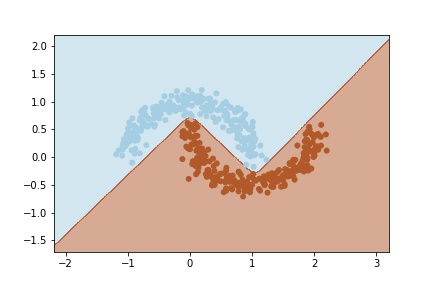}
         \caption{clean test, balanced CT} \label{fig:TOy a)}
     \end{subfigure}
      &
     \begin{subfigure}{0.225\textwidth}
         \includegraphics[width=1.2\linewidth]{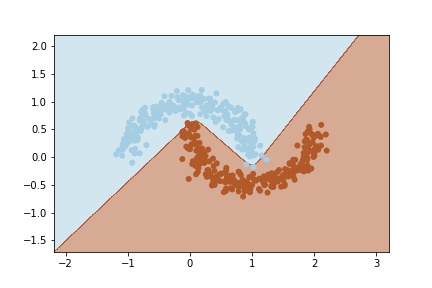}
         \caption{clean test, balanced AT} \label{fig:TOy b)}
     \end{subfigure}
      &
     \begin{subfigure}{0.223\textwidth}
         \includegraphics[width=1.2\linewidth]{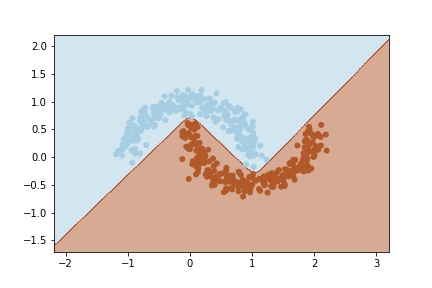}
         \caption{clean test, imbalanced CT} \label{fig:TOy c)}
     \end{subfigure}
      &
     \begin{subfigure}{0.223\textwidth}
         \includegraphics[width=1.2\linewidth]{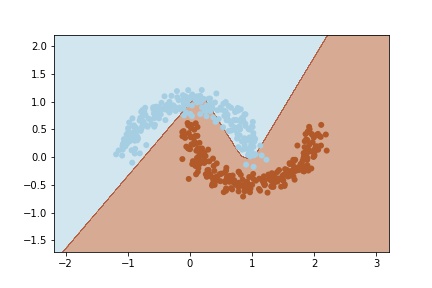}
         \caption{clean test, imbalanced AT} \label{fig:TOy d)}
     \end{subfigure}
       \\
     \begin{subfigure}{0.223\textwidth}
         \includegraphics[width=1.2\linewidth]{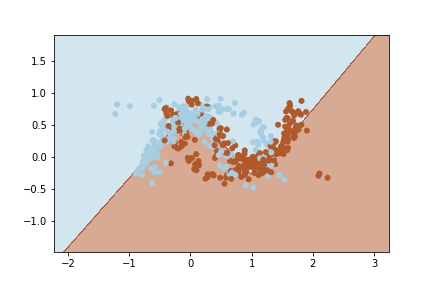}
         \caption{RT test, balanced CT} \label{fig:TOy e)}
     \end{subfigure}
      &
     \begin{subfigure}{0.223\textwidth}
         \includegraphics[width=1.2\linewidth]{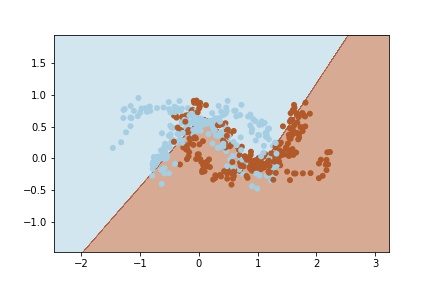}
         \caption{RT test, balanced AT} \label{fig:TOy f)}
     \end{subfigure}
      &
     \begin{subfigure}{0.223\textwidth}
         \includegraphics[width=1.2\linewidth]{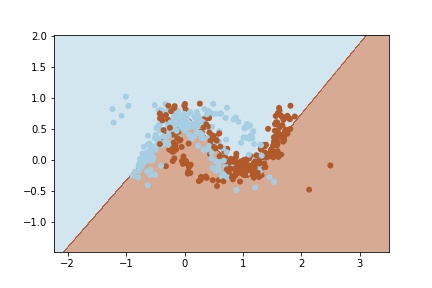}
         \caption{RT test, imbalanced CT} \label{fig:TOy g)}
     \end{subfigure}
      &
     \begin{subfigure}{0.223\textwidth}
         \includegraphics[width=1.2\linewidth]{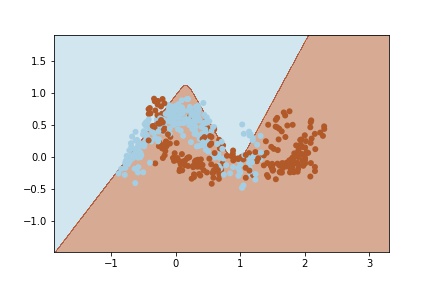}
         \caption{RT test, imbalanced AT} \label{fig:TOy h)}
     \end{subfigure}
     \\
\end{tabular}

\caption{Decision boundary for training data and test sample distribution on the toy task. Clean and robustness test samples are dotted with decision boundary obtained by AT on balanced and imbalanced clean training data. RT means robustness.}
\label{fig:Toy}
\end{figure*}

\paragraph{Attack Overwhelming by Class-Imbalance}
Class-imbalance of CICL increases the number of adversarial attacks of a majority class, which overwhelms the number of clean examples for the minority class.
The class-imbalance, which is a well-known but still unsolved problem of CICL, causes the imbalance of adversarial attacks, because AT generates them by distorting all clean samples in a mini-batch. 
For example, AT using 10\% of training data for a class, and 90\% for the others exactly inherits the rate to the generated adversarial examples~\cite{goodfellow2014explaining}. 
In usual CICL settings~\cite{buzzega2020dark}, the class-imbalance is a common property, and therefore the imbalanced attack occurs in most CICLs to naively adopt AT.


\paragraph{Weak Resistance to Inbound Attacks by Class-Imbalance}
The small clean examples by class-imbalance weakens the resistance to inbound attacks. 
We use the term, the inbound attack for a class, to indicate closely located adversarial examples from the other classes. 
When the inbound attacks are trained in AT, the number of clean examples has an important role of resisting to distortion of existing information by the attacks in the model. 
The resistance is weaken for the minority class in CICL, which has insufficient clean examples compared to the other majority classes. For example, in rehearsal-based method, model can access full current task data but only access previous task data within a very small memory size compared to the current task size. This imbalance of previous-current tasks ratio gap as CL progresses.

\paragraph{Problem: Decision Boundary Distortion}
The two properties, attack overwhelming of the majority class and weakening resistance of the minority class, cause critical distortion of trained information from clean data. 
This phenomenon appears as the distortion of decision boundaries. 
The overwhelming attacks increase the inbound attacks for the minority classes, and the minority classes has insufficient resistance to the attack by lacking clean examples. 
Combination of these two properties then increases the gap of training loss for clean examples and adversarial examples, and then distorts the decision boundary obtained by clean examples.
The distortion directly causes accuracy drop because the clean test samples follows the imbalance rate of clean training examples to build the decision boundary. 
In the case of robustness, the test samples are balanced adversarial attacks in AT, which builds different decision boundary to the distorted one based on the imbalanced attacks.
This difference causes robustness errors.

\begin{table}[h]
    \centering
    \begin{tabular}{lrr}
    \hline
    training type & accuracy(\%) & robustness(\%) \\ \hline
    Balanced CT & 100.0 & 43.6 \\
    Balanced AT & 99.2 & 69.6 \\
    Imbalanced CT & 100.0 & 46.2 \\ 
    Imbalanced AT & 93.4 & 52.6\\
    \hline
    \end{tabular}
    \caption{Numerical results of accuracy and robustness on the toy task.}
    \label{tab:toy_result}
\end{table}
\paragraph{Settings for Empirical Analysis}
We prepared a toy binary classification task to preliminarily verify the distortion phenomenon. In the task, we generated the same number of crescent-shaped input representations for each of two classes as Figure~\ref{fig:Toy}, like~\cite{altinisik2022a3t}. Each class has 1000 input samples.

\begin{figure*}[ht]

  \centering
    \fbox{\includegraphics[width=\textwidth]{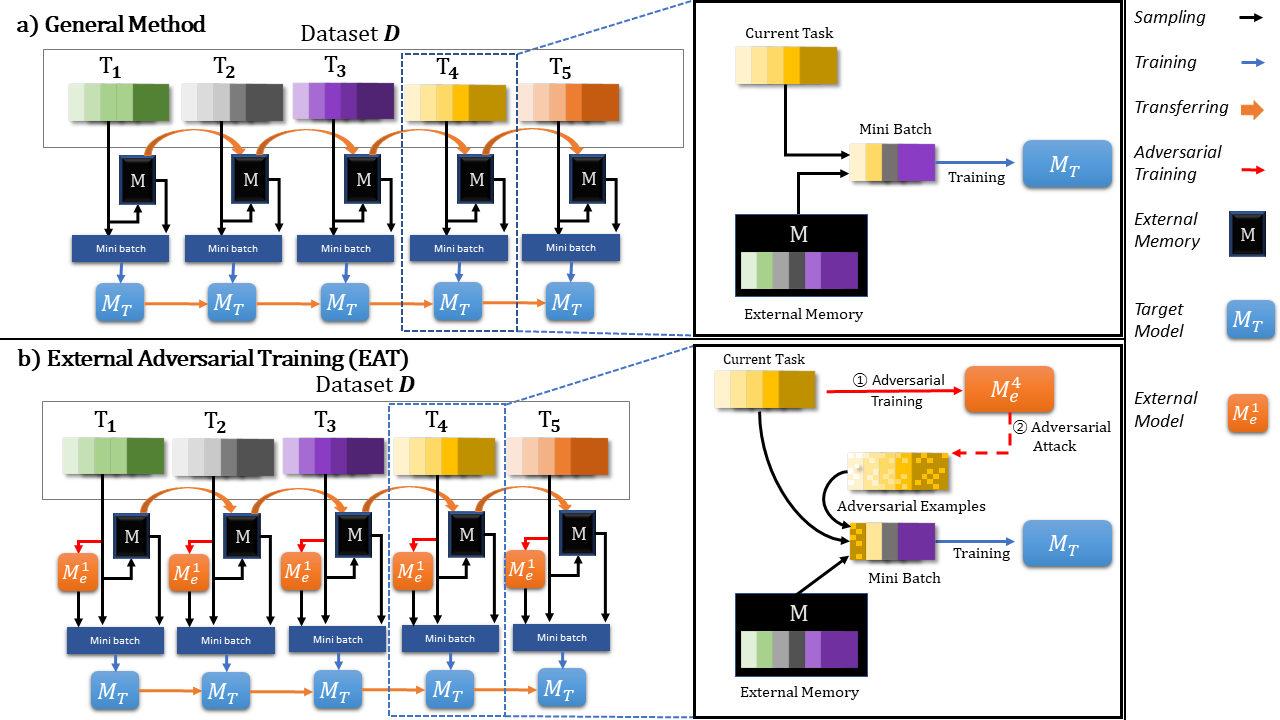}}
  \caption{Overview of a) general experience replay and b) proposed EAT method on it in CICL. ($M_T$: the target model transferred over time steps)}
\label{fig:overview}
\end{figure*}

We trained a simple linear network which composed three hidden nodes, two layer feed-forward network on the data in the four conditions of training data: 1) balanced clean data, 2) balanced clean data with balanced adversarial examples, 3) imbalanced clean data, and 4) imbalanced clean data with imbalanced adversarial examples (1:9). Training used SGD optimizer, learning rate as 0.1, doing 500 epochs. For adversarial training, using PGD attack with 10 iters.

The trained models are used for plotting their decision boundaries by generating predicted classes over representation space as shown in Figure~\ref{fig:Toy}.
The boundary is tested on balanced clean samples and balanced adversarial examples that shown as dot distribution in the first and second row of the figure. Detail accuracy can be seen at Table~\ref{tab:toy_result}.


\paragraph{Distortion Compared to Clean Test}
In Figure~\ref{fig:TOy a)}, the model trained with balanced clean data shows clear decision boundary to distinguish the clean test samples. 
In Figure~\ref{fig:TOy b)}, the model which trained with balanced adversarial samples changes the boundary slightly, but still maintains the boundary of clean data. 
Using the imbalanced adversarial examples (in Figure~\ref{fig:TOy d)}), the model largely moves the boundary from the majority (red) to the minority of classes (blue) and incorrectly classify more blue test samples.
The results imply that the imbalanced adversarial training has a potential to distort the boundary and destroy the original decision boundary built by clean data. Note that there has been no critical clean accuracy degradation in imbalanced clean training (in Figure~\ref{fig:TOy c)}). This degradation in performance and increase poor robustness occur only when AT is combined in the imbalanced setting. It does not happen in simple imbalanced setting.

\paragraph{Distortion Compared to Robustness Test}
In Figure~\ref{fig:TOy d)}, balanced clean training shows the base robustness to the adversarial attacks generated for its trained model. 
Applying AT to the balanced data, Figure~\ref{fig:TOy e)}, the trained model shows significantly improved robustness, which is a desirable gain by AT in an ordinary balanced training environment. 
However, imbalanced AT shows less improvement, compared to the imbalanced clean training. 
In the balanced case, the boundary is nearly changed, but the imbalance case shows the shifted boundary toward the blue area when AT is applied.
Then, most of robustness test samples for blue class are incorrectly classified.
This result also provides an evidence of robustness degradation by decision boundary distortion.


\section{Method}

\paragraph{Simple Solution: External Adversarial Training}
In Figure~\ref{fig:overview}, the details of EAT to CICL with experience replay setting are shown.
Compared to typical AT, EAT creates an additional external model whose backbone has the same network architecture to the CL model shared over tasks (Target model).
At each step, the method creates an external model, trains it via AT only for the current task at the step from the scratch, generates adversarial examples, and deleted.
Then, Target model trains with current task data, replayed samples from memory, and the generated adversarial samples without AT.  
Detail process is described in Algorithm~\ref{alg:eat}.
Note that EAT doesn't need any extra external memory size. External model deleted after generate adversarial examples, do not saved for future tasks.

\begin{algorithm}[h!]
\begin{algorithmic}[1]
\State Given a task stream $D=\{T_1,\cdots,T_N\}$, a target model $M_T$, and an external memory \textrm{M}, external model $M_{e}$
\Procedure{ER+EAT}{$D$, $M_b$, $M$}
\For{$i$ from 1 to N}
\State $AE_i \leftarrow$ EAT ($M_{e}$, $T_i$) 
\For{$(x,y)$ in $T_i\cup AE_i$}
\If{i $>$ 1}
\State random sampling ($x',y'$) from \textrm{M} 
\EndIf
\State training $M_T$ on $(x',y') \cup (x, y)$
\EndFor
\State update \textrm{M} by \textit{Reservoir sampling}
\EndFor
\State \textbf{return} $M_T$
\EndProcedure
\State
\Procedure{EAT}{$M_{T}$, $T_i$}
\State adversarial training $M_{e}^i$ on $T_i$
\State generation of $AE_i$ from $T_i$ via $M_{e}^i$
\State \textbf{return} $AE_i$
\EndProcedure
\end{algorithmic}
\caption{External Adversarial Training in ER}
\label{alg:eat}
\end{algorithm}

\begin{figure}[h!]
  \centering
    \includegraphics[width=0.45\textwidth]{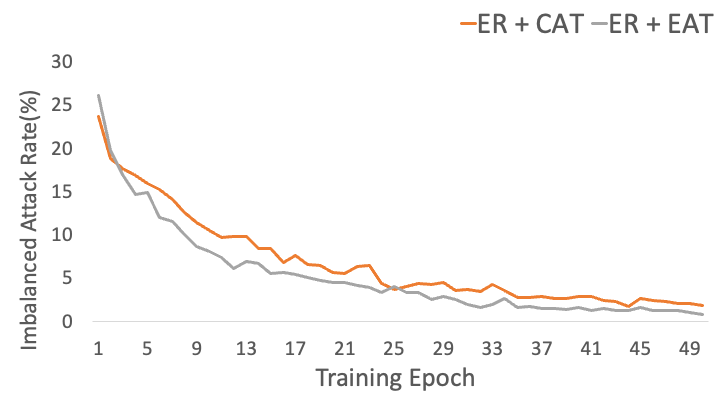}
  \caption{Rate of adversarial attacks from current tasks to previous tasks by training epochs}
  \label{heatmap}
\label{fig:heatmap}
\end{figure}

\paragraph{Motivation: Effective Exclusion of AT on Class-Imbalance}
Motivation of EAT is to effectively exclude imbalanced AT for reducing the distortion effect.
In CICL, the imbalanced AT appears by the imbalanced size of current task data and replayed samples, and therefore AT over only different tasks suffers from the distortion problem. 
Excluding the cases of applying AT over different tasks is a practically achievable way for the goal, because the class-imbalance is a nature of CICL method, which has no clear solution in a limited computing environment. 
A simple way of the exclusion is to learn Target model only on the current task data, called current task adversarial training (CAT) in this paper. 
However, this method generates attacks from a current task to other different tasks in CICL settings to incrementally expand the class set for prediction. 
To enhance the exclusion, EAT uses an external model focused on attacks between classes of a current task. 
In Figure~\ref{fig:heatmap}, the rate of adversarial samples between different Tasks is shown. This experiment is conducted on split CIFAR-10 and the other settings are shown in Section~\ref{setting}. 
In the result, EAT shows higher rate than CAT over all training epochs, which verifies more effective exclusion of EAT. 
In fact, the unclear exclusion of CAT improves largely decreases the accuracy and robustness slightly as shown in  Figure~\ref{fig:drawback_AT}.

\section{Experiment}
\label{setting}
\paragraph{Model and Training}
Similar to~\cite{chaudhry2019tiny,lopez2017gradient,mai2022online}, we set ResNet18~\cite{he2016deep} as the target model. We set hyperparameters same as~\cite{buzzega2020dark} which chosen by grid-search on a 10\% validation set. Every model using SGD optimizer. we set the number of epochs to 50 for each task as~\cite{buzzega2020dark,ho2020contrastive}.

\paragraph{Memory Update}
After training of the base model at a epochs, the external memory is updated by inserting samples randomly selected from the task at the step. This memory update method is called as \textit{Reservoir sampling}. If memory is already full, we randomly choose data in the memory and replace this data to new data.

\paragraph{Datasets}
We use three datasets, Split-CIFAR-10, Split-CIFAR-100, and Split-MiniImageNet. Each set is created through splitting original data by classes, composing of classes for each task, and ordering the tasks as a stream.  The task composition and ordering determine the information for transfer over tasks, and their different settings cause the large change of results. For clear analysis, we fixed task composition in ascending order of labels.

\begin{table}[h]

    \setlength{\tabcolsep}{0.7\tabcolsep}
    \centering
    \begin{tabular}{lcc}
        \hline
        Datasets & Split CIFAR-10 & Split MiniImageNet \\ \hline
        task &5 & 20\\
        classes / task & 2 & 5 \\
        tr. samples / task & 2000 & 500 \\
        te. sample / task &10 &100 \\
        image Size &32x32 & 84x84\\
        \hline
    \end{tabular}%
    \label{tab:Dataset}
    \caption{Statistics of CICL benchmarks. (tr: training, te: test)}
\end{table}




\paragraph{Method Setting} 
We use adversarial attack method for EAT as PGD attack~\cite{madry2017towards}. To adversarialy training the external model, we perform adversarial training in 10 epoch. For PGD attack, we use 4 iterations. For FGSM attack and PGD attack, we used $\alpha$ of 0.0078, and $\epsilon$ of 0.0314. For Comparison, we test adversarial training using FGSM and PGD attck. The setup of PGD and FGSM attack used for adversarial training is same as EAT. To test robustness, we used PGD attack with 4 iterations. EAT was applied to ER, DER, and DERpp, which are methods that can be applied without deformation. We compared our method with knowledge distillation method (iCaRL~\cite{rebuffi2017icarl}), and 7 rehearsal-based methods (ER, GEM~\cite{lopez2017gradient}, FDR~\cite{benjamin2018measuring}, HAL~\cite{chaudhry2021using}, DER, DERpp). Like~\cite{buzzega2020dark}, we do not compared with HAL and GEM in seq Tiny-ImageNet setting because its untractable time.

\section{Results and Discussion}

\begin{table*}[t!]
    \resizebox{\textwidth}{!}{
    \begin{tabular}{llrrrrrrr}
    \hline
    \multicolumn{1}{c}{Buffer} & \multicolumn{1}{|c|}{Method} & \multicolumn{3}{c|}{CIFAR-10}                           & \multicolumn{3}{c}{Tiny ImageNet} \\ \cline{3-8}
\multicolumn{1}{c}{Size}                             & \multicolumn{1}{|l|}{}           & Accuracy(\%)     & Robustness(\%) & \multicolumn{1}{r|}{Accuracy*(\%)}   &
Accuracy(\%)   & Robustness(\%) & Accuracy*(\%)   \\ \hline
                                                 & \multicolumn{1}{|l|}{GEM~\cite{lopez2017gradient}}        & 29.75 (±3.1)      & 11.44 (±0.2)   & \multicolumn{1}{r|}{25.54 (±3.2)}     & --             & --             & --  \\
                                                 & \multicolumn{1}{|l|}{iCaRL~\cite{rebuffi2017icarl}}      & 54.70 (±0.3)      & 12.72 (±1.5)    & \multicolumn{1}{r|}{49.02 (±3.2)}     & 9.03 (±0.4)     & 1.20 (±0.2)     & 7.53 (±0.8)  \\
                                                 & \multicolumn{1}{|l|}{FDR~\cite{benjamin2018measuring}}        & 30.78 (±8.3)      & 10.02 (±2.1)    & \multicolumn{1}{r|}{30.91 (±2.7)}     & 8.96 (±0.2)     & 0.89 (±0.4)     & 8.70 (±0.2)  \\
                                                 & \multicolumn{1}{|l|}{HAL~\cite{chaudhry2018efficient}}        & 37.26 (±1.2)      & 11.66 (±0.3)    & \multicolumn{1}{r|}{32.36 (±2.7)}     & --             & --             & --  \\
                                                 \cline{2-8}
                                                 & \multicolumn{1}{|l|}{ER~\cite{chaudhry2019tiny}}         & 49.50 (±1.5)       & 13.15 (±0.7)    & \multicolumn{1}{r|}{44.79 (±1.9)}     & 8.74 (±0.2)     & 1.11 (±0.3)     & 8.49 (±0.2)  \\
\multicolumn{1}{c}{\rotatebox[origin=c]{0}{200}}                          & \multicolumn{1}{|l|}{ER + AT}    & 45.52 (±0.5)      & 18.39 (±0.6)    & \multicolumn{1}{r|}{-- }             & 8.49 (±0.2)     & 2.72 (±0.4)     & --  \\
                                                 & \multicolumn{1}{|l|}{ER + EAT}   & 50.16 (±0.4)      & 20.08 (±1.0)    & \multicolumn{1}{r|}{--}              & 8.62 (±0.4)     & 3.69 (±0.2)     & --  \\                                                                                                  \cline{2-8}
                                                 \cline{2-8}
                                                 & \multicolumn{1}{|l|}{DER~\cite{buzzega2020dark}}        & 60.02 (±1.7)      & 15.47 (±1.2)    & \multicolumn{1}{r|}{61.23 (±1.1)}     & 11.84 (±1.2)    & 1.33 (±0.2)     & 11.87 (±0.8)  \\
                                                 & \multicolumn{1}{|l|}{DER + AT}   & 40.62 (±3.4)     & 20.38 (±0.6)    & \multicolumn{1}{r|}{--}              & 8.23 (±1.0)     & 1.43 (±0.3)             & --  \\
                                                 & \multicolumn{1}{|l|}{DER + EAT}  & 52.21 (±3.8)     & 21.39 (±1.2)    & \multicolumn{1}{r|}{--}              & 8.70 (±0.7)     & 2.32 (±0.2)             & --  \\
                                                 \cline{2-8}
                                                 & \multicolumn{1}{|l|}{DERpp~\cite{buzzega2020dark}}      & 62.57 (±1.6)      & 16.48 (±1.0)    & \multicolumn{1}{r|}{64.88 (±1.2)}     & 11.83 (±0.6)    & 1.46 (±0.1)     & 10.96 (±1.2)  \\
                                                 & \multicolumn{1}{|l|}{DERpp + AT} & 50.51 (±2.2)      & 23.49 (±3.4)    & \multicolumn{1}{r|}{--}              & 8.12 (±0.7)     & 1.97 (±0.3)            & --  \\
                                                 & \multicolumn{1}{|l|}{DERpp + EAT}& 57.78 (±2.7)      & 24.00 (±1.3)    & \multicolumn{1}{r|}{--}              & 9.04 (±1.0)     & 2.88 (±0.3)     & --  \\ \hline
                                                 & \multicolumn{1}{|l|}{GEM~\cite{lopez2017gradient}}        & 26.98 (±1.9)      & 11.32 (±0.5)    & \multicolumn{1}{r|}{26.20 (±1.3)}    & --             & --             & --          \\
                                                 & \multicolumn{1}{|l|}{iCaRL~\cite{rebuffi2017icarl}}      & 53.80 (±0.7)     & 13.79 (±1.6)    & \multicolumn{1}{r|}{47.55 (±4.0)}    & 10.90 (±1.2)    & 1.82 (±0.2)     & 9.38 (±1.5)  \\
                                                 & \multicolumn{1}{|l|}{FDR~\cite{benjamin2018measuring}}        & 24.03 (±2.1)      & 12.27 (±0.5)    & \multicolumn{1}{r|}{28.71 (±3.2)}    & 12.72 (±0.5)    & 1.06 (±0.4)     & 10.54 (±0.2) \\
                                                 & \multicolumn{1}{|l|}{HAL~\cite{chaudhry2018efficient}}        & 48.94 (±0.6)      & 15.91 (±0.8)    & \multicolumn{1}{r|}{41.79 (±4.5)}    & --             & --             & --          \\
                                                  \cline{2-8}
                                                 & \multicolumn{1}{|l|}{ER~\cite{chaudhry2019tiny}}         & 61.49 (±1.8)      & 15.48 (±1.5)    & \multicolumn{1}{r|}{57.74 (±0.3)}    & 11.86 (±2.8)    & 1.13 (±0.1).    & 9.99 (±0.2)  \\
\multicolumn{1}{c}{\rotatebox[origin=c]{0}{500}}                          & \multicolumn{1}{|l|}{ER + AT}    & 58.18 (±1.1)      & 16.68 (±0.5)    & \multicolumn{1}{r|}{--}              & 9.40 (±1.0)      & 3.75 (±0.3)     & --          \\
                                                 & \multicolumn{1}{|l|}{ER + EAT}   & 62.76 (±1.6)      & 23.43 (±2.1)     & \multicolumn{1}{r|}{--}              & 9.52 (±0.9)     & 4.06 (±0.3)     & --          \\
                                                 \cline{2-8}
                                                 & \multicolumn{1}{|l|}{DER~\cite{buzzega2020dark}}        & 72.29 (±0.9)      & 18.20 (±0.5)    & \multicolumn{1}{r|}{70.51 (±1.7)}    & 16.20 (±0.7)    & 1.60 (±0.3)     & 17.75 (±1.1) \\
                                                 & \multicolumn{1}{|l|}{DER + AT}   & 54.82 (±1.1)      & 18.18 (±1.0)    & \multicolumn{1}{r|}{--}              & 8.96 (±0.7)     & 2.13 (±0.8)     & --          \\
                                                 & \multicolumn{1}{|l|}{DER + EAT}  & 60.10 (±2.7)       & 18.70 (±1.2)    & \multicolumn{1}{r|}{--}              & 10.87 (±1.1)    & 2.66 (±0.2)     & --          \\
                                                 \cline{2-8}
                                                 & \multicolumn{1}{|l|}{DERpp~\cite{buzzega2020dark}}      & 73.50 (±1.3)       & 17.21 (±0.5)    & \multicolumn{1}{r|}{72.70 (±1.4)}    & 17.21 (±0.5)    & 1.75 (±0.3)     & 19.38 (±1.4) \\
                                                 & \multicolumn{1}{|l|}{DERpp + AT} & 61.59 (±0.3)      & 22.27 (±0.9)    & \multicolumn{1}{r|}{--}              & 11.21 (±1.0)    & 2.98 (±0.3)     & --          \\
                                                 & \multicolumn{1}{|l|}{DERpp + EAT}& 64.10 (±2.7)       & 23.11 (±2.1)    & \multicolumn{1}{r|}{--}              & 13.53 (±0.8)    & 4.41 (±0.5)     & --          \\ \hline

    \end{tabular}}
    \caption{Accuracy and robustness on split CIFAR-10 and split Tiny Imagenet dataset. Accuracy* is the reference result. Every value is averaged over 3 trial.}

\label{Tag:All}

\end{table*}


%

\paragraph{Performance Comparison with State-of-The-Art}
The performance of accuracy and robustness in some of state-of-the-art models are shown in Table~\ref{Tag:All}. 
The results are categorized to two cases using 200 and 500 buffer size for experience replay. 
In each memory setting, we reproduce state-of-the-art methods and their results are close to their reference accuracy results with some variance.

In the accuracy results, AT significantly decreases accuracy of experience replay methods in all cases compared to their original accuracy, whereas EAT show significantly larger accuracy than AT. 
In the robustness results, AT improves robustness of all base methods. 
EAT significantly increases robustness again and shows the best value over all methods in the table. 

The results imply that EAT effectively solves the problems on accuracy and robustness drop of AT in CICL. 
Furthermore, EAT is the most effective method to enhance robustness in CICL. 
Less accuracy than the best original method is the trade-off between accuracy and robustness, which is the property of AT observed in usual cases. 

\paragraph{Robustness and Accuracy on Each Task After Training}
\begin{figure}[h!]
  \centering
    \includegraphics[width=0.45\textwidth]{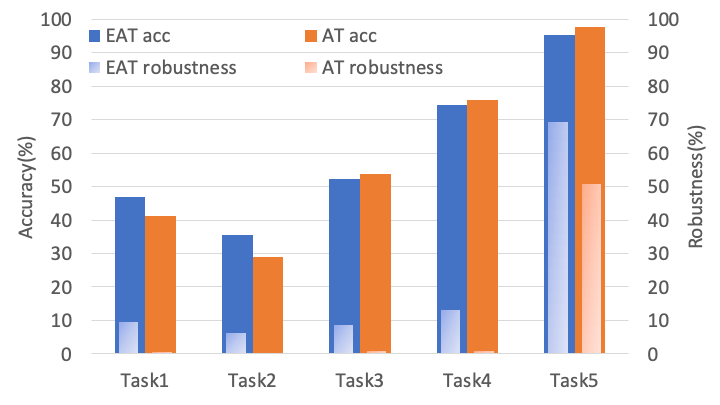}
  \caption{Accuracy and robustness of each task at the final step of ER + AT and ER + EAT (acc: accuracy)}
\label{fig:FINAL}
\end{figure}
 Figure~\ref{fig:FINAL} shows detail robustness of AT and EAT for each task after training over all tasks on CIFAR-10. 
 In the results, EAT shows higher robustness than AT in all tasks. AT nearly improves any previous tasks except current task (Task5). 
 The accuracy of EAT is higher at Task1 and Task2, which are the oldest two tasks, whereas AT shows slightly higher accuracy in the recent tasks.
 Considering the total accuracy and robustness increase of EAT, the results imply that EAT improves both, specifically improves older accuracy better, and significantly improves all robustness. 
Note that EAT has never learned inter-task adversarial attacks, but the robustness increases overall tasks. 
This is the strong evidence of drawbacks of unnecessary class-imbalance attacks of AT between tasks . 


%

\paragraph{Performance Difference over Time Steps}
\begin{figure}[h]

  \centering
    \includegraphics[width=0.45\textwidth]{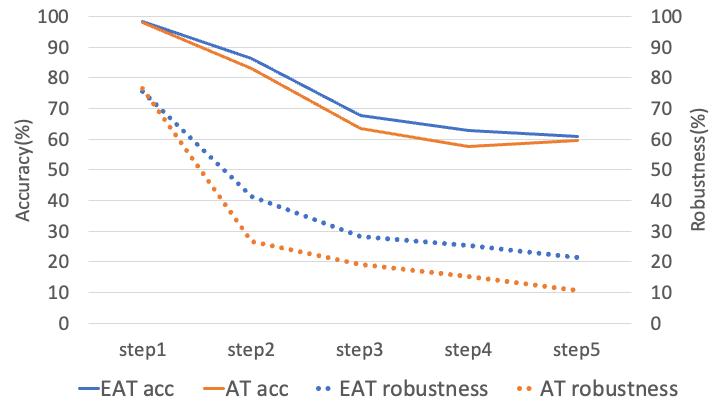}
  \caption{Accuracy and robustness of ER + AT and ER + EAT at each steps  (acc: accuracy)}
  \label{fig:steps}
\end{figure}
Figure~\ref{fig:steps} shows accuracy and robustness averaged over involved tasks at each step in training. 
The accuracy gradually decreases by steps in CICL settings, while the model is repeatedly trained for a new task and forgets the previous task information. This phenomenon of CICL appears for both AT and EAT, but their overall accuracy is slightly higher with EAT. 
The robustness is similar, but not exactly equal at step1, which is caused by randomness of adversarial attacks of AT.
The difference of the robustness results significantly increases at step2 and it shows the similar robustness until step5. 
As step2 includes only Task1 and Task2 but step5 includes all Tasks, the remaining difference imply that CICL settings with AT have sufficiently large robustness degradation when adding a new task. 




\paragraph{Reducing Computational Cost}
Both EAT and AT are computationally expensive to build and train adversarial samples. 
In particular, EAT is more expensive because it trains and uses new external models.
For practical use, the cost may be a limitation, so we also verify the performance of EAT in better CICL settings to use faster attack method, FGSM~\cite{szegedy2013intriguing}.
Compared to 4-PGD attack, the method reduces the time complexity to about 25\%~\cite{szegedy2013intriguing}. 
Table~\ref{table:FGSM} shows the performance in the efficient setting.
In the results, accuracy and robustness are still improved by EAT significantly, so the limit of EAT in computational cost can be sufficiently alleviated.

\begin{table}[h!]
\centering
\begin{tabular}{lcc}
\hline
Method                & Accuracy & Robustness \\
\hline
ER              & 61.49     & 15.48      \\
ER + AT (FGSM)  & 59.28     & 15.80            \\
ER + EAT (FGSM) & 62.43     & 18.21          \\
\hline
\end{tabular}

\caption{Accuracy and Robustness of AT and EAT using computationally efficient FGSM attack method}
\label{table:FGSM}
\end{table}




\section{Conclusion}
In this paper, we show that existing AT do not work well in class-incremental continual learning setting with experience replay. We argue that its cause lies in AT on the class-imbalance data and its distortion of decision boundaries results in accuracy and robustness drop. To solve the distortion, we introduced EAT that effectively excludes the imbalanced AT between different tasks.
In the experiments on CICL benchmarks, we verify that our method significantly improves both accuracy compared to AT suffering from the negative effect of class-imbalance. 
Moreover, EAT provides the new state-of-the-art defence performance (robustness) in CICL with ER environment. 
 
\section{Future Works}
Although robustness of several methods has been investigated in this paper, the robustness of many CL methods is still insufficient. In addition, there is also a lack of study about how adversarial defense method except adversarial training affect to CL. Wide and various study on the adversarial robustness in CL need to be studied with future work. To the best of our knowledge, this study is the first to study adversarial defenses specialized in CL. Affordable and effective adversarial defenses specialized in CL should also be studied in the future.

\section{Related Work}
\paragraph{Continual learning}
CL can be divided into several categories according to problem setting and constraints. One group extends the architecture of the model for each new task. Another approach is to regularize the model with respect to previous task knowledge while training new tasks. And Rehearsal method use stored data or samples from generative models to resist catastrophic forgetting. Rehearsal method is very effective in class-incremental CL, but there are additional computational cost and memory costs. Recent rehearsal-free method have shown high performance with little memory cost using vision transformer and prompt tuning. This setting is more realistic and shows higher performance than setting starting with scratch. In this paper, we focus on the setting of class-incremental CL from the scratch. 

\paragraph{Adversarial Defense}

There are various adversarial defense methods that have been proposed in the literature, including adversarial training, defensive distillation, input preprocessing methods, and model ensemble methods. Defensive distillation~\cite{papernot2016distillation} is another method that improves the robustness of the model by distilling the knowledge from a robust model into a less robust one. Input preprocessing~\cite{dziugaite2016study} methods aim to preprocess the input data to remove the adversarial perturbations before feeding it to the model. Model ensemble methods~\cite{pang2019improving}, on the other hand, aim to increase the robustness by combining the predictions of multiple models.

Other methods such as gradient masking, Randomized smoothing and Adversarial Detection are also proposed in recent years. Gradient masking~\cite{lee2020gradient} is a method that hides the gradients of the model to prevent the gradient-based attacks. Randomized smoothing~\cite{cohen2019certified} is a method that makes the model more robust by adding random noise to the input data. Adversarial Detection~\cite{liu2018adversarial} is a method that aims to detect the adversarial examples and discard them before they are fed to the model.

\paragraph{Continual learning with adversarial defense}
Efforts to incorporate adversarial robustness into CL have not been long studied. \cite{9892970} is studied on how to increase robustness in joint training using continual pruning method. But this study didn't study about how to increase robustness in CL. \cite{9937385} using the robust and non-robust data set found in~\cite{ilyas2019adversarial} to increase clean accuracy of continual learning model. They also did experiments on robustness of CL model, but only conducted experiments in seq CIFAR-10 with large memory (=16000). And their goal is to increase clean accuracy, they have not studied how to increase adversarial robustness. In this paper, we first studied how to increase adversarial robustness in CL, and measured robustness of various method in various setting.
\section{Acknowledgement} 
This work was partially supported by the National Research Foundation of Korea (NRF) grant funded by the Korea government (MSIT) (2022R1A2C2012054) and by Culture, Sports and Tourism R\&D Program through the Korea Creative Content Agency grant funded by the Ministry of Culture, Sports and Tourism in 2022 (Project Name: Development of service robot and contents supporting children's reading activities based on artificial intelligence Project Number:R2022060001, Contribution Rate: 50\%).

\appendix
\section{}

\bibliographystyle{named}
\bibliography{ijcai23}

\end{document}


\maketitle

\appendix
\section{Detail Setting of Toy problem}
We used a model that performs binary classification by receiving two dimensional data as input. The model has two linear layers and the number of hidden nodes is 3. We fixed the running rate to 0.1 and the training epoch to 500. The data used is 1000 pieces per label, forming a crescent-shaped distribution. The attack used for adversarial training and robustness measurements was PGD attack, with number of iteration of 10, and alpha and eps values of 0.1.

\section{Detail setting of Figure 1}
We used setting same as main experiment setting as paragraph 5 for ER, ER+AT and ER+CAT. For joint training and joint adversarial training, we used 200 epoch for training whole dataset. For CAT, we only perform adversarial training on current task data. For AT, we perform adversarial training on both current task data and memory retrieved data.

\section{Detail setting of Figure 4}
We used setting same as main experiment setting as paragraph 5. We investigated the process when the model learns the second task. We investigated the rate at which adversarial examples used in adversarial training are predicted by previous tasks. Since neither CAT nor EAT proceeds to attack with memory data, all adversarial examples have the current task as a real label. We investigate the proportion of cases where the real label is the current task, but the label confused by adversarial attacks is the previous task.

\section{Detail setting of reproduce}
We created Example.ipynb, an example code for reproducibility. This file can basically implement the reproduction of ER+EAT on split CIFAR-10. Various factors in this file can be adjusted to reproduce the experiments presented in this paper. A detailed reproduction method is shown in Example.ipynb.